\documentclass[10pt,twocolumn,letterpaper]{article}
    
    \usepackage{cvpr}
    \usepackage{times}
    \usepackage{epsfig}
    \usepackage{graphicx}
    \usepackage{amsmath}
    \usepackage{amssymb}
    \usepackage{enumitem}
    \usepackage{multirow}
    \usepackage{csquotes}
    \usepackage{flushend}
    \usepackage{mathtools, cuted}
    \usepackage{appendix}
    
    \usepackage[pagebackref=true,breaklinks=true,letterpaper=true,colorlinks,bookmarks=false]{hyperref}

    \graphicspath{fig}
    
    \usepackage{color}

    \cvprfinalcopy 
    
    \def\cvprPaperID{674} 

    \ifcvprfinal\fi
\begin{document}

    \title{Polysemous Visual-Semantic Embedding for Cross-Modal Retrieval\thanks{Code and data: \url{https://yalesong.github.io/pvse}}}
    \author{Yale Song\\
    Microsoft Cognition\\
    {\tt\small yalesong@microsoft.com}
    \and
    Mohammad Soleymani\\
    USC Institute for Creative Technologies\\
    {\tt\small soleymani@ict.usc.edu}
    }
    
    \maketitle
    
    \begin{abstract}
    Visual-semantic embedding aims to find a shared latent space where related visual and textual instances are close to each other. Most current methods learn injective embedding functions that map an instance to a single point in the shared space. Unfortunately, injective embedding cannot effectively handle polysemous instances with multiple possible meanings; at best, it would find an average representation of different meanings. This hinders its use in real-world scenarios where individual instances and their cross-modal associations are often ambiguous. In this work, we introduce Polysemous Instance Embedding Networks (PIE-Nets) that compute multiple and diverse representations of an instance by combining global context with locally-guided features via multi-head self-attention and residual learning. To learn visual-semantic embedding, we tie-up two PIE-Nets and optimize them jointly in the multiple instance learning framework. Most existing work on cross-modal retrieval focuses on image-text data. Here, we also tackle a more challenging case of video-text retrieval. To facilitate further research in video-text retrieval, we release a new dataset of 50K video-sentence pairs collected from social media, dubbed MRW (my reaction when). We demonstrate our approach on both image-text and video-text retrieval scenarios using MS-COCO, TGIF, and our new MRW dataset.
    \end{abstract}

    \section{Introduction}
    \label{sec:introduction}
    Visual-semantic embedding~\cite{frome-nips13,karpathy-cvpr15} aims to find a joint mapping of instances from visual and textual domains to a shared embedding space so that related instances from source domains are mapped to nearby places in the target space. This has a variety of downstream applications in computer vision including tagging~\cite{frome-nips13}, retrieval~\cite{gong-ijcv14}, captioning~\cite{karpathy-cvpr15}, visual question answering~\cite{jang-cvpr17}. 
    
    Formally, the goal of visual-semantic embedding is to learn two mapping functions $f: \mathcal{X} \rightarrow \mathcal{Z}$ and $g: \mathcal{Y} \rightarrow \mathcal{Z}$ jointly, where $\mathcal{X}$ and $\mathcal{Y}$ are visual and textual domains, respectively, and $\mathcal{Z}$ is a shared embedding space. The functions are often designed to be \textit{injective} so that there is a one-to-one mapping from an instance $x$ (or $y$) to a single point $z \in \mathbb{R}^d$ in the embedding space. They are often optimized to satisfy the following constraint:
    \begin{equation}
    \label{eq:sil-objective}
        d( f(x_i), g(y_i) ) < d( f(x_i), g(y_j) ), \:\:\:\: \forall i \neq j  
    \end{equation}
    where $d(\cdot,\cdot)$ is a certain distance measure, such as Euclidean and cosine distance. This simple and intuitive setup, which we refer to as \textit{injective instance embedding}, is currently the most popular approach in the literature~\cite{wang-arxiv16}.
    
    \begin{figure}
       \centering
       \includegraphics[width=\linewidth]{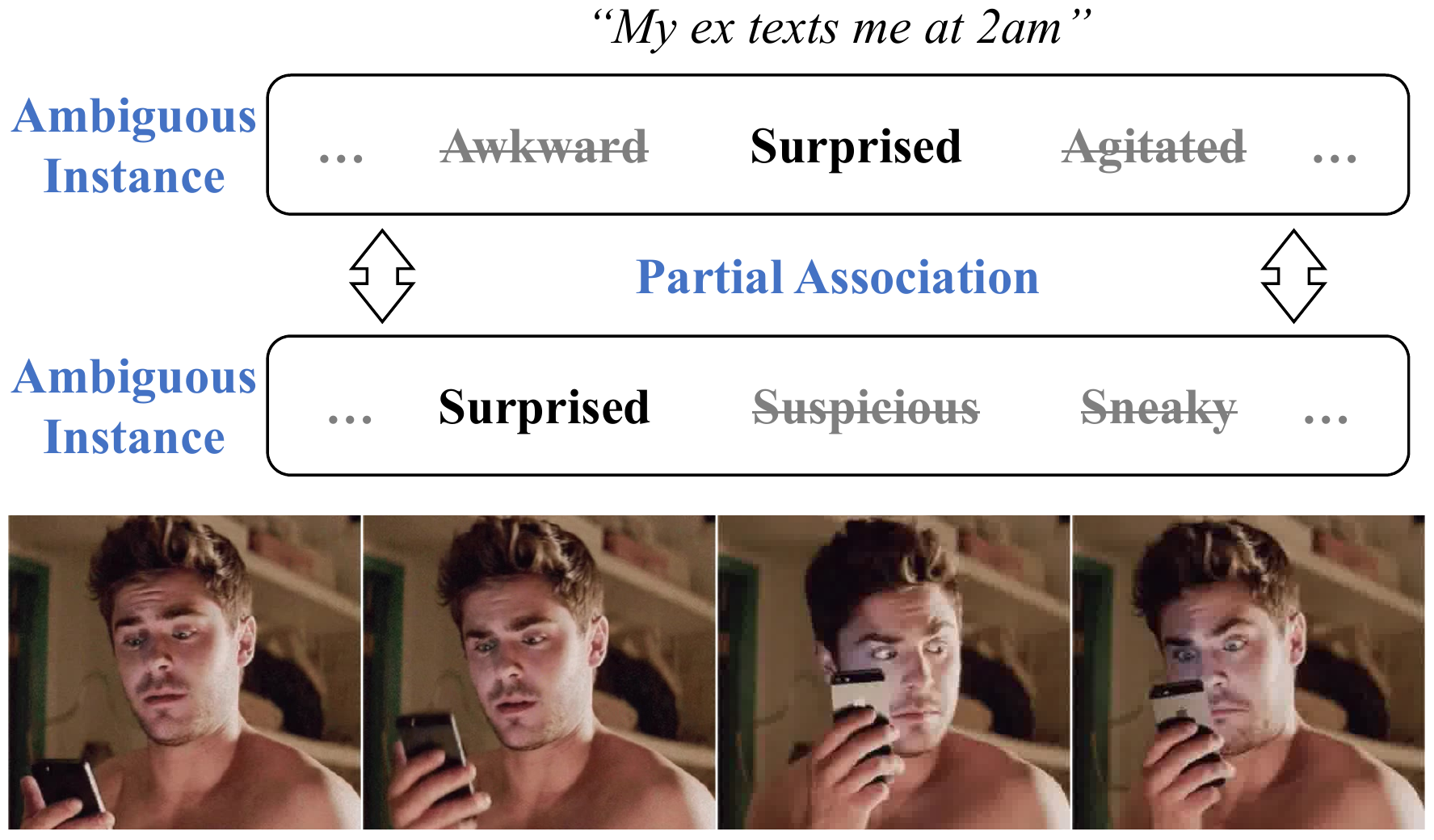}
       \caption{Cross-modal retrieval in the real-world could be challenging with \textit{ambiguous instances} (each instance can have multiple meanings/concepts) and their \textit{partial associations} (not all individual meanings/concepts may match). Addressing these two challenges is the focus of this work.
       }
       \label{fig:concept}
    \end{figure}
    
    Unfortunately, injective embedding can suffer when there is \textit{ambiguity} in individual instances. Consider an ambiguous instance with multiple meanings/senses, e.g., polysemy words and images containing multiple objects. Even though each of the meanings/senses can map to different points in the embedding space, injective embedding is always forced to find a single point, which could be an (inaccurate) weighted geometric mean of all the desirable points. The issue gets intensified for videos and sentences because the ambiguity in individual images and words can aggregate and get compounded, severely limiting its use in real-world applications such as text-to-video retrieval.
    
    Another case where injective embedding could be problematic is \textit{partial} cross-domain association, a characteristic commonly observed in the real-world datasets. For instance, a text sentence may describe only certain regions of an image while ignoring other parts~\cite{xu-icml15}, and a video may contain extra frames not described by its associated sentence~\cite{li-cvpr16}. These associations are implicit/hidden, making it unclear which part(s) of the image/video the text description refers to. This is especially problematic for injective embedding because information about any ignored parts will be lost in the mapped point and, once mapped, there is no way to recover from the information loss.
    
    In this work, we address the above issues by (1) formulating instance embedding as a one-to-many mapping task and (2) optimizing the mapping functions to be robust to ambiguous instances and partial cross-modal associations. 
    
    To address the issues with ambiguous instances, we propose a novel one-to-many instance embedding model, \textbf{P}olysemous \textbf{I}nstance \textbf{E}mbedding \textbf{Net}work (PIE-Net), which extracts $K$ embeddings of each instance by combining global and local information of its input. Specifically, we obtain $K$ \textit{locally-guided} representations by attending to different parts of an input instance (e.g., regions, frames, words) using a multi-head self-attention module~\cite{lin-iclr17,vaswani-nips17}. We then combine each of such local representation with global representation via residual learning~\cite{he-cvpr16} to avoid learning redundant information. Furthermore, to prevent the $K$ embeddings from collapsing into the mode (or the mean) of all the desirable embeddings, we regularize the $K$ locally-guided representations to be diverse. To our knowledge, we are the first to apply multi-head self-attention with residual learning for the application of instance embedding.
    
\begin{figure}
       \centering
       \includegraphics[width=.8\linewidth]{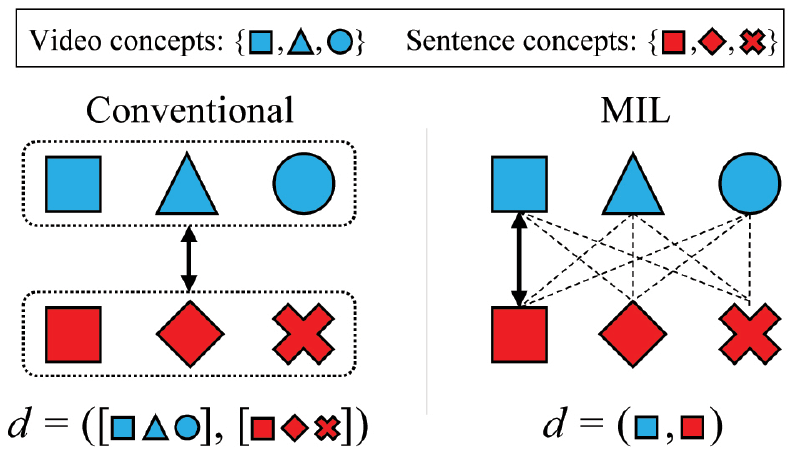}
       \vspace{2pt}
       \caption{We represent each instance with $k$ embeddings, each representing different parts of the instance, e.g., regions of an image, frames of a video, or words of a sentence. Conventional approaches measure the visual-semantic distance by considering all $k$ embeddings, and thus would suffer when not all concepts are related. We instead assume there is a partial match and measure the distance between only the most related combination (squares).}
       \label{fig:mil-concept}
    \end{figure}

    To address the partial association issue, we tie-up two PIE-Nets and train our model in the multiple-instance learning (MIL) framework~\cite{dietterich-ai97}. We call this approach \textbf{P}olysemous \textbf{V}isual-\textbf{S}emantic \textbf{E}mbedding (PVSE). Our intuition is: when two instances are only partially associated, the learning constraint of Equation~\eqref{eq:sil-objective} will unnecessarily penalize embedding mismatches because it expects two instances to be perfectly associated. Capitalizing on our one-to-many instance embedding, our MIL objective relaxes the constraint of Equation~\eqref{eq:sil-objective} so that only one of $K \times K$ embedding pairs is well-aligned, making our model more robust to partial cross-domain association. We illustrate this intuition in Figure~\ref{fig:mil-concept}. This relaxation, however, could cause a discrepancy between two embedding distributions because $(K \times K - 1)$ embedding pairs are left unconstrained. We thus regularize the learned embedding space by minimizing the discrepancy using the Maximum Mean Discrepancy (MMD)~\cite{gretton-nips07}, a popular technique for determining whether two sets of data are from the same probability distribution. 
    
    We demonstrate our approach on two cross-modal retrieval scenarios: image-text and video-text. For image-text retrieval, we evaluate on the MS-COCO dataset~\cite{lin-eccv14}; for video-text retrieval, we evaluate on the TGIF dataset~\cite{li-cvpr16} as well as our new MRW (my reaction when) dataset, which we collected to promote further research in cross-modal video-text retrieval under ambiguity and partial association. The dataset contains 50K video-sentence pairs collected from social media, where the videos depict physical or emotional reactions to certain situations described in text. We compare our method with well-established baselines and carefully conduct an ablation study to justify various design choices. We report strong performance on all three datasets, and achieve the state-of-the-art result on image-to-text retrieval task on the MS-COCO dataset.
    
    
    \section{Related Work} 
    \label{sec:related_work}
    
    Here we briefly review some of the most relevant work on instance embedding for cross-modal retrieval.
    
    \textbf{Correlation maximization:} Most existing methods are based on one-to-one mapping of instances into a shared embedding space. One popular approach is maximizing correlation between related instances in the embedding space. Rasiwasia~\etal~\cite{rasiwasia-mm10} use canonical correlation analysis (CCA) to maximize correlation between images and text, while Gong~\etal~\cite{gong-ijcv14} extend CCA to a triplet scenario, e.g., images, tags, and their semantic concepts. Most recent methods incorporate deep neural networks to learn their embedding models in an end-to-end fashion. Andrew~\etal~\cite{andrew-icml13} propose deep CCA (DCCA), and Yan~\etal~\cite{yan-cvpr15} apply it to image-to-sentence and sentence-to-image retrieval.
    
    \textbf{Triplet ranking:} Another popular approach is based on triplet ranking~\cite{frome-nips13,kiros-arxiv14,wang-cvpr16,ye-eccv18}, which encourages the distance between positive pairs (e.g., ground-truth pairs) to be closer than negative pairs (e.g., randomly selected pairs). Frome~\etal~\cite{frome-nips13} propose a deep visual-semantic embedding (DeViSE) model, using a hinge loss to implement triplet ranking. Faghri~\etal~\cite{faghri-bmvc17} extend this with the idea of hard negative mining, which focuses on maximum violating negative pairs, and report improved convergence rates. 
    
    \begin{figure*}
       \centering
       \includegraphics[width=1\linewidth]{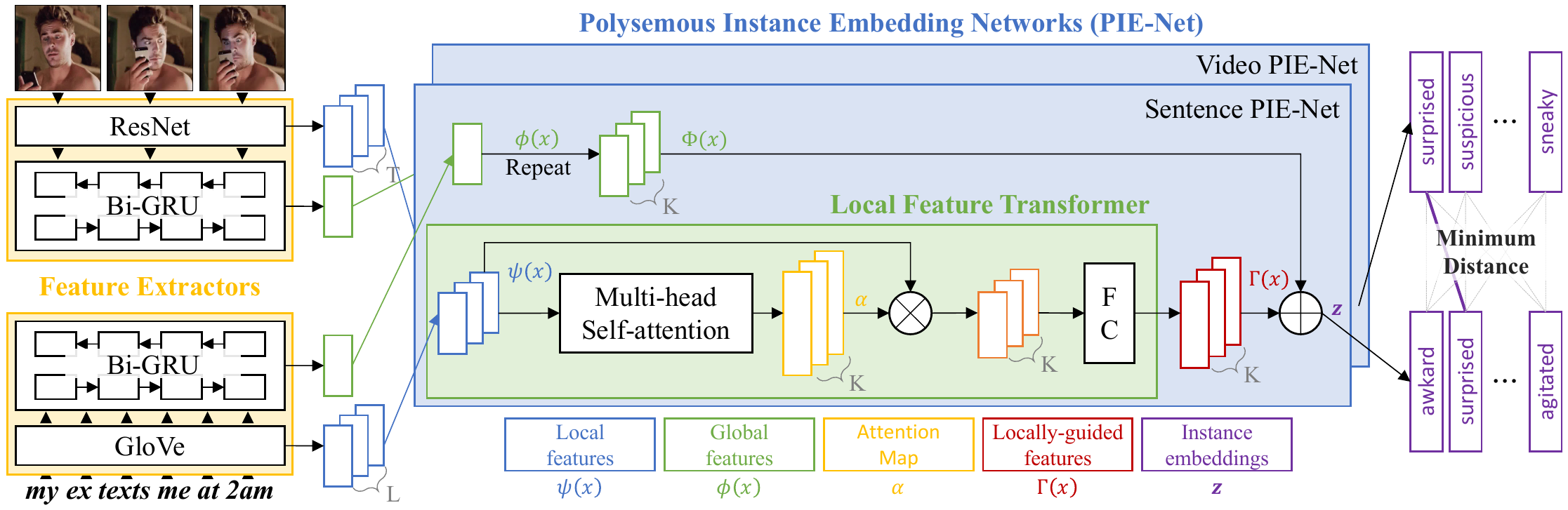}
       \caption{The architecture of Polysemous Visual-Semantic Embedding (PVSE) for video-sentence data.}
       \label{fig:pvse}
    \end{figure*}
    
    \textbf{Learning with auxiliary tasks:} Several methods learn the embeddings in conjunction by solving auxiliary tasks, e.g., signal reconstruction~\cite{feng-mm14,eisenschtat-cvpr17,tsai-iccv17}, semantic concept categorization~\cite{rasiwasia-mm10,huang-cvpr18}, and minimizing the divergence between embedding distributions induced by different modalities~\cite{tsai-iccv17,zhang-eccv18}. Adversarial training~\cite{goodfellow-nips14} is also used by many: Wang~\etal~\cite{wang-mm17} encourage the embeddings from different modalities to be indistinguishable using a domain discriminator, while Gu~\etal~\cite{gu-cvpr18} learn the embeddings with image-to-text and text-to-image synthesis tasks in the adversarial learning framework.
    
    \textbf{Attention-based embedding:} All the above approaches are based on one-to-one mapping and thus could suffer from polysemous instances. To alleviate this, recent methods incorporate cross-attention mechanisms to selectively attend to local parts of an instance \textit{given the context of} a conditioning instance from another modality~\cite{huang-cvpr17,lee-eccv18}, e.g., attend to different image regions given different text queries. Intuitively, this can resolve the issues with ambiguous instances and their partial associations because the same instance can be mapped to different points depending on the presence of the conditioning instance. However, such approach comes with computational overhead at inference time because each query instance needs to be encoded as many times as the number of references instances in the database; this severely limits its use in real-world applications. Different from previous approaches, our method is based on multi-head self-attention~\cite{lin-iclr17,vaswani-nips17} which does not require a conditioning instance when encoding, and therefore each instance is encoded only once, significantly reducing computational overhead at inference time.
    
    \textbf{Beyond injective embedding:} Similar to our motivation, some attempts have been made to go beyond the injective mapping. One approach is to design the embedding function to be stochastic and map an instance to a certain probability distribution (e.g., Gaussian) instead of a single point~\cite{ren-mm16,mukherjee-emnlp16,oh-arxiv18}. However, learning distributions is typically difficult/expensive and often lead to approximate solutions such as Monte Carlo sampling. 
    
    The work most similar to ours is by Ren~\etal~\cite{ren-bmvc17}, where they compute multiple representations of an image by extracting local features using the region proposal method~\cite{girshick-iccv15}; text instances are still represented by a single embedding vector. Different from theirs, our method computes multiple and diverse representations from both modalities, where each representation is a combination of global context and locally-guided features, instead of just a local feature. Song~\etal~\cite{song-arxiv18}, a prequel to this work, also compute multiple representations of each instance using multi-head self-attention. We extend their approach by combining global and locally-guided features via residual learning. We also extend the preliminary version of the MRW dataset with an increased number of sample pairs. Lastly, we report more comprehensive experimental results, adding results on the MS-COCO~\cite{lin-eccv14} dataset for image-text cross-retrieval.

    \section{Approach}
    \label{sec:approach}
    
    Our Polysemous Visual-Semantic Embedding (PVSE) model, shown in Figure~\ref{fig:pvse}, is composed of modality-specific feature extractors followed by two sub-networks with an identical architecture; we call the sub-network Polysemous Instance Embedding Network (PIE-Net). The two PIE-Nets are independent of each other and do not share the weights.
    
    The PIE-Net takes as input a global context vector and multiple local feature vectors (Section~\ref{subsec:encoders}), computes locally-guided features using the local feature transformer (Section~\ref{subsec:transformer}), and outputs $K$ embeddings by combining the global context vector with locally-guided features (Section~\ref{subsec:fusion}). We train the PVSE model in the Multiple Instance Learning (MIL)~\cite{dietterich-ai97} framework. We explain how we make our model robust to ambiguous instances and partial cross-modal associations via our loss functions (Section~\ref{subsec:optimization}) and finish with implementation details (Section~\ref{subsec:details}).
    

    \subsection{Modality-Specific Feature Encoder}
    \label{subsec:encoders}
    
    
    \textbf{Image encoder:} We use the ResNet-152~\cite{he-cvpr16} pretrained on ImageNet~\cite{russakovsky-ijcv15} to encode an image $x$. We take the feature map before the final average pooling layer as local features $\Psi(x) \in \mathbb{R}^{7 \times 7 \times 2048}$. We then apply average pooling to $\Psi(x)$ and feed the output to one fully-connected layer to obtain global features $\phi(x) \in \mathbb{R}^{H}$. 
    
    \textbf{Video encoder:} We use the ResNet-152 to encode each of $T$ frames from a video $x$, taking the 2048-dim output from the final average pooling layer, and use them as local features $\Psi(x) \in \mathbb{R}^{T \times 2048}$. We then feed $\Psi(x)$ into a bidirectional GRU (bi-GRU)~\cite{cho-emnlp14} with $H$ hidden units, and take the final hidden states as global features $\phi(x) \in \mathbb{R}^{H}$.
    
    \textbf{Sentence encoder:} We encode each of $L$ words from a sentence $x$ using the GloVe~\cite{pennington-emnlp14} pretrained on the CommonCrawl dataset, producing $L$ 300-dim vectors, and use them as local features $\Psi(x) \in \mathbb{R}^{L \times 300}$. We then feed them into a bi-GRU with $H$ hidden units, and take the final hidden states as global features $\phi(x) \in \mathbb{R}^{H}$.

    \subsection{Local Feature Transformer}
    \label{subsec:transformer}
    
    The local feature transformer takes local features $\Psi(x)$ and transforms them into $K$ locally-guided representations $\Upsilon(x)$. Our intuition is that different combinations of local information could yield diverse and refined representations of an instance. We implement this intuition by employing a multi-head self-attention module to obtain $K$ attention maps, prepare $K$ combinations of local features by attending to different parts of an instance, and apply non-linear transformations to obtain $K$ locally-guided representations.
    
    We use a two-layer perceptron to implement the multi-head self-attention module.\footnote{We have experimented with a more sophisticated version of the multi-head self-attention~\cite{vaswani-nips17}, but it did not improve performance further.} Given local features $\Psi(x) \in \mathbb{R}^{B \times D}$\footnote{$B$ is 49 ($=7 \times 7$) for images, $T$ for videos, and $L$ for sentences; $D$ is 2048 for images and videos, and 300 for sentences}, it computes $K$ attention maps $\alpha \in \mathbb{R}^{K \times B}$: 
    \begin{equation}
        \alpha = \mbox{softmax}\left( w_2 \: \mbox{tanh} \left( w_1 \Psi(x)^{\intercal} \right)\right)
    \label{eq:attn1}
    \end{equation}
    where $w_2 \in \mathbb{R}^{K \times A}$, $w_1 \in \mathbb{R}^{A \times D}$; we set $A=D/2$ per empirical evidence. The softmax is applied row-wise so that each of the $K$ attention coefficients sum up to one. 
    
    Finally, we multiply the attention map with local features and further apply a non-linear transformation to obtain $K$ locally-guided representations $\Upsilon(x) \in \mathbb{R}^{K \times H}$:
    \begin{equation}
        \Upsilon(x) = \sigma\left( (\alpha \Psi(x)) w_3 + b_3 \right)
    \label{eq:attn2}
    \end{equation}
    where $w_3 \in \mathbb{R}^{D \times H}$ and $b_3 \in \mathbb{R}^{H}$. We use the sigmoid as our activation function $\sigma(\cdot)$.

    \subsection{Feature Fusion With Residual Learning}
    \label{subsec:fusion}
    
    The fusion block combines global features $\phi(x)$ and locally-guided features $\Upsilon(x)$ to obtain the final $K$ embedding output. We note that there is an inherent information overlap between the two features (both are derived from the same instance). To prevent $\Upsilon(x)$ from becoming redundant with $\phi(x)$ and encourage it to learn only locally-specific information, we cast the feature fusion as a residual learning task. Specifically, we consider $\phi(x)$ as input to the residual block and $\Upsilon(x)$ as residuals with its own parameters to optimize ($w_1, w_2, w_3, b_3$). As shown in \cite{he-cvpr16}, this residual mapping makes it easier to optimize the parameters associated with $\Upsilon(x)$, helping us find meaningful locally-specific information; in the extreme case, if global features $\phi(x)$ were the optimal, the residuals will be pushed to zero and the approach will fall back to the standard injective embedding.
    
    We compute $K$ embedding vectors $z \in \mathbb{R}^{K \times H}$ as:
    \begin{equation}
        z = \mbox{LayerNorm} \left( \Phi(x) + \Upsilon(x) \right)
    \label{eq:final_embedding}
    \end{equation}
    where $\Phi(x) \in \mathbb{R}^{K \times H}$ is $K$ repetitions of $\phi(x)$. Following \cite{vaswani-nips17}, we apply the layer normalization~\cite{ba-arxiv16} to the output.

    \subsection{Optimization and Inference}
    \label{subsec:optimization}
    
    Given a dataset $\mathcal{D}=\{ (x_i, y_i)\}_{i=1}^{N}$ with $N$ instance pairs ($x$ are either images or videos, $y$ are sentences), we optimize our PVSE model to minimize a learning objective:
    \begin{equation}
        \mathcal{L} = \mathcal{L}_{mil} + \lambda_1 \mathcal{L}_{div} + \lambda_2 \mathcal{L}_{mmd}
    \label{eq:our_objective}
    \end{equation}
    where $\lambda_1$ and $\lambda_2$ are scalar weights that balance the influence of the loss terms. We describe each loss term below.

    \textbf{MIL Loss:} 
    We train our model in the Multiple Instance Learning (MIL) framework~\cite{dietterich-ai97}, designing a learning constraint for the cross-modal retrieval scenario:
    \begin{equation}
        \min_{p,q} d( z^x_{i,p}, z^y_{i,q} ) < d( z^x_{i,p}, z^y_{j,q} ), \:\:\:\: \forall i \neq j, \:\: \forall p, q
    \label{eq:mil-objective}
    \end{equation}
    where $z^x$ and $z^y$ are the PIE-Net embeddings of $x$ and $y$, respectively, and $p,q = 1, \cdots, K$. We use the cosine distance as our distance metric, $d(a,b) = (a \cdot b) / (\|a\|\|b\|$).
    
    Making an analogy to the MIL for binary classification~\cite{amores-ai13}, the left side of the constraint is the ``positive'' bag where at least one of $K \times K$ embedding pairs is assumed to be positive (match), while the right side is the ``negative'' bag containing only negative (mismatch) pairs. Optimizing under this constraint allows our model to be robust to partial cross-modal association because it can ignore mismatching embedding pairs of partially associated instances.
    
    We implement the above constraint by designing our MIL loss function $\mathcal{L}_{mil}$ to be:
    \begin{equation}
        \frac{1}{N^2} \sum_{i,j}^{N} 
            \max\left(0, 
                \rho - \min_{p,q} d(z^x_{i,p}, z^y_{j,q}) + \min_{p,q} d(z^x_{i,p}, z^y_{i,q})
            \right) \nonumber
    \end{equation}
    where $\rho$ is a margin parameter. Notice that we have the min operator for $d(z^x_{i,p}, z^y_{j,q})$, similar to \cite{ren-bmvc17}; this can be seen as a form of hard negative mining, which we found to be effective and accelerate the convergence. 
    
    \begin{figure*}
       \centering
       \includegraphics[width=0.95\linewidth]{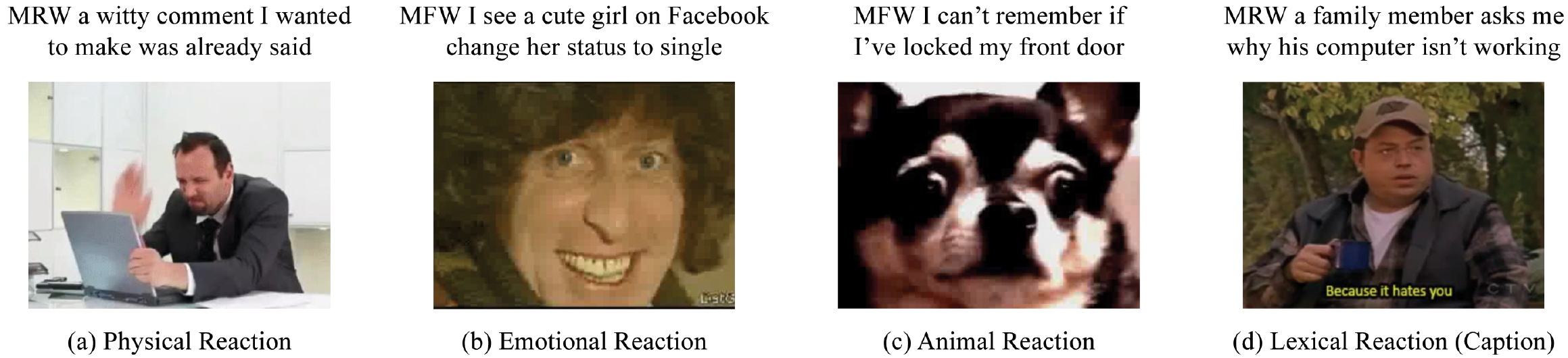}
       \caption{Our dataset contains videos depicting \textit{reactions} to the situations described in the corresponding sentences. Here we show the four most common reaction types: (a) physical, (b) emotional, (c) animal, (d) lexical.}
       \label{fig:mrw_examples}
    \end{figure*}

    \textbf{Diversity Loss:} 
    To ensure that our PIE-Net produces diverse representations of an instance, we design a diversity loss $\mathcal{L}_{div}$ that penalizes the redundancy among $K$ locally-guided features. To measure the redundancy, we compute a Gram matrix of $\Upsilon(x)$ (and of $\Upsilon(y)$) that encodes the correlations between all combinations of locally-guided features, i.e., $G_{i,j} = \sum_{h} \Upsilon(x)_{ih} \Upsilon(x)_{jh}$. We normalize each $\Upsilon(x)_{i}$ prior to the computation so that they are on an $l_2$ ball. 
    
    The diagonal entries in $G$ are always one (they are on a unit ball); the off-diagonals are zero iff two locally-guided features are orthogonal to each other. Therefore, the sum of off-diagonal entries in $G$ indicates the redundancy among $K$ locally-guided features. Based on this, we define our diversity loss as:
    \begin{equation}
        \mathcal{L}_{div} = 
        \frac{1}{K^2} \left( \|G^x - I\|_2 + \|G^y - I\|_2 \right)
    \end{equation}
    where $G^x$ and $G^y$ are the gram matrices of $\Upsilon(x)$ and $\Upsilon(y)$, respectively, and $I \in \mathbb{R}^{K \times K}$ is an identity matrix.
    
    Note that we do not compute the diversity loss on the final embedding representations $z^x$ and $z^y$ because they already have global information baked in, making the orthogonality constraint invalid. This also ensures that the loss gets back-propagated through appropriate parts in the computational graph, and does not affect the global feature encoders, i.e., the FC layer for the image encoder, and the bi-GRUs for the video and sentence encoders.
    
    

    \textbf{Domain Discrepancy Loss:}
    Optimizing our model under the MIL loss has one drawback: two distributions induced by $z^x$ and $z^y$, which we denote by $Z^x$ and $Z^y$, respectively, may diverge quickly because we only consider the minimum distance pair, $\min_{p,q} d(z^x_{p}, z^y_{q})$, in loss computation and let the other $(K \times K - 1)$ pairs left to be unconstrained. It is therefore necessary to regularize the discrepancy between the two distributions.
    
    One popular way to measure the discrepancy between two probability distributions is the Maximum Mean Discrepancy (MMD)~\cite{gretton-nips07}. The MMD between two distributions $P$ and $Q$ over a function space $\mathcal{F}$ is 
    \begin{equation}
        \mbox{MMD}(P,Q) = \sup_{f \in \mathcal{F}} \left( 
            \mathbb{E}_{X \sim P} \left[ f(X) \right]
            - \mathbb{E}_{Y \sim Q} \left[ f(Y) \right]
        \right)  
    \label{eq:mmd}
    \end{equation}
    When $\mathcal{F}$ is a reproducing kernel Hilbert space (RKHS) with a kernel $\kappa: \mathcal{X} \times \mathcal{X} \rightarrow \mathbb{R}$ that measures the similarity between two samples, Gretton~\etal~\cite{gretton-nips07} showed that the supremum is achieved at $f(x) = \mathbb{E}_{X' \sim P}[\kappa(x, X')] - \mathbb{E}_{X' \sim Q}[\kappa(x, X')]$. Substituting this to Equation~\eqref{eq:mmd} and squaring the result, and approximating the expectation over our empirical distributions $Z^x$ and $Z^y$, we have our domain discrepancy loss $\mathcal{L}_{mmd}$ defined as
    \begin{equation}
        \frac{ 
            \sum \kappa(z^x_{i,p}, z^x_{j,q}) 
            - 2\sum \kappa(z^x_{i,p}, z^y_{j,q})
            + \sum \kappa(z^y_{i,p}, z^y_{j,q})
        }{K^2N^2} \nonumber
    \end{equation}
    where the summation in each term is taken over all pairs of embeddings $(i,j,p,q) \in [1, \cdots, K^2N^2]$. We use a radial basis function (RBF) kernel as our kernel function.
    
    \textbf{Inference:} At test time, we assume a database of $M$ instances (e.g., videos) and their $KM$ embedding vectors. Given a query instance (e.g., a sentence), we compute $K$ embedding vectors and find the best matching instance in the database by comparing the cosine distances between all $K^2M$ combinations of embeddings.
    
    \subsection{Implementation Details}
    \label{subsec:details}
    
    We subsample frames at 8 FPS and store them in a binary storage format.\footnote{\url{https://github.com/TwentyBN/GulpIO}} We set the maximum length of video to be 8 frames; for videos longer than 8 frames we select random subsequences during training, while during inference we sample 8 frames evenly spread across each video. We do not limit the sentence length as it has a minimal effect on the GPU memory footprint. We cross-validate the optimal hyper-parameter settings, varying $K \in [1:8]$, $H \in [512, 1024, 2048], \rho \in [0.1:1.0], \lambda_1, \lambda_2 \in [0.1, 0.01, 0.001]$. We use the AMSGRAD optimizer~\cite{reddi-iclr18} with an initial learning rate of 2e-4 and reduce it by half when the loss stagnates. We train our model end-to-end, except for the pretrained CNN weights, for 50 epochs with a batch of 128 samples. We then finetune the whole model (including the CNN weights) for another 50 epochs. 

    \section{MRW Dataset}
    \label{sec:dataset}
    To promote future research in video-text cross-modal retrieval, especially with ambiguous instances and their partial cross-domain association, we release a new dataset of 50K video-sentence pairs collected from social media; we call our dataset MRW (my reaction when).
    
    Table~\ref{tab:datasets} provides descriptive statistics of several video-sentence datasets. Most existing datasets are designed for video captioning~\cite{rohrbach-ijcv17,xu-cvpr16,li-cvpr16}, with sentences providing textual descriptions of visual content in videos (video $\to$ text relationship). Our dataset is unique in that it provides videos that display physical or emotional reactions to the given sentences (text $\to$ video relationship); these are called \textit{reaction GIFs}. According to a subreddit \texttt{r/reactiongif}\footnote{\url{https://www.reddit.com/r/reactiongifs}}:
    \begin{displayquote}
    \textit{\small A reaction GIF is a physical or emotional response that is captured in an animated GIF which you can link in response to someone or something on the Internet. The reaction must not be in response to something that happens within the GIF, or it is considered a ``scene''.}
    \end{displayquote}
    This definition clearly differentiates ours from existing datasets: There is an inherently weaker association of concepts between video and text; see Figure~\ref{fig:mrw_examples}. This introduces several additional challenges to cross-modal retrieval, part of which are the focus of this work, i.e., dealing with ambiguous instances and partial cross-domain association. We provide detailed data analyses and compare it with existing video captioning datasets in the supplementary material.
    
    \begin{table}[t]
    \footnotesize
    \centering
    \begin{tabular}{l|rrrc}
     & \#clips & \#sentences & vocab & text source\\ \hline 
    LSMDC16~\cite{rohrbach-ijcv17} & 128,085 & 128,085 & 22,898 & DVS \\
    MSR-VTT~\cite{xu-cvpr16} & 10,000 & 200,000 & 29,316 & AMT\\
    TGIF~\cite{li-cvpr16} & 100,000 & 125,781 & 11,806 & AMT\\
    DiDeMo~\cite{hendricks-iccv17} & 26,982 & 40,543 & 7,785& AMT\\ \hline
    \textbf{MRW} & 50,107 & 50,107 & 34,835 & In-the-wild
    \end{tabular}
    \vspace{5pt}
    \caption{\textbf{Descriptive statistics} of our dataset compared to existing video-sentence datasets. }
    \label{tab:datasets}
    \end{table}
    
    \begin{table*}[t]
    \centering
    \begin{tabular}{l|ccc|ccc|ccc|ccc}
    \hline
    \multicolumn{1}{c|}{\multirow{3}{*}{Method}} 
        & \multicolumn{6}{c|}{1K Test Images} 
        & \multicolumn{6}{c}{5K Test Images} \\ \cline{2-13}
        & \multicolumn{3}{c|}{Image-to-Text} 
        & \multicolumn{3}{c|}{Text-to-Image} 
        & \multicolumn{3}{c|}{Image-to-Text} 
        & \multicolumn{3}{c}{Text-to-Image} \\ 
        & R@1 & R@5 & R@10 & R@1 & R@5 & R@10 
        & R@1 & R@5 & R@10 & R@1 & R@5 & R@10 \\ \hline
    DVSA~\cite{karpathy-cvpr15} 
        & 38.4 & 69.9 & 80.5 & 27.4 & 60.2 & 74.8 & 16.5 & 39.2 & 52.0 & 10.7 & 29.6 & 42.2 \\
    GMM-FV~\cite{klein-cvpr15} 
        & 39.4 & 67.9 & 80.9 & 25.1 & 59.8 & 76.6 & 17.3 & 39.0 & 50.2 & 10.8 & 28.3 & 40.1 \\
    m-CNN~\cite{ma-iccv15}
        & 42.8 & 73.1 & 84.1 & 32.6 & 68.6 & 82.8 & - & - & - & - & - & - \\
    Order~\cite{vendrov-iclr16}
        & 46.7 & - & 88.9 & 37.9 & - & 85.9 & 23.3 & - & 65.0 & 18.0 & - & 57.6 \\
    DSPE~\cite{wang-cvpr16}
        & 50.1 & 79.7 & 89.2 & 39.6 & 75.2 & 86.9 & - & - & - & - & - & - \\
    VQA-A~\cite{lin-eccv16}
        & 50.5 & 80.1 & 89.7 & 37.0 & 70.9 & 82.9 & 23.5 & 50.7 & 63.6 & 16.7 & 40.5 & 53.8 \\
    2WayNet~\cite{eisenschtat-cvpr17}
        & 55.8 & 75.2 & - & 39.7 & 63.3 & - & - & - & - & - & - & - \\
    RRF-Net~\cite{liu-iccv17}
        & 56.4 & 85.3 & 91.5 & 43.9 & 78.1 & 88.6 & - & - & - & - & - & - \\
    CMPM~\cite{zhang-eccv18}
        & 56.1 & 86.3 & 92.9 & 44.6 & 78.8 & 89.0 & 31.1 & 60.7 & 73.9 & 22.9 & 50.2 & 63.8 \\
    VSE++~\cite{faghri-bmvc17}  
        & 64.6 & 90.0 & 95.7& 52.0 & 84.3 & 92.0 & 41.3 & 71.1 & 81.2 & 30.3 & 59.4 & 72.4 \\
    GXN~\cite{gu-cvpr18}
        & 68.5 & - & \textbf{97.9} & \underline{56.6} & - & \underline{94.5} & - & - & - & - & - & - \\
    SCO~\cite{huang-cvpr18}
        & \textbf{69.9} & \textbf{92.9} & \underline{97.5} & \textbf{56.7} & \textbf{87.5} & \textbf{94.8} & \underline{42.8} & 72.3 & 83.0 & \textbf{33.1} & \underline{62.9} & \textbf{75.5} \\
    \hline 
    PVSE (K=1) 
        & 66.7 & 91.0 & 96.2 & 53.5 & 85.1 & 92.7 & 41.7 & \underline{73.0} & \underline{83.0} & 30.6 & 61.4 & 73.6 \\
    \textbf{PVSE}  
        & \underline{69.2} & \underline{91.6} & 96.6 & 55.2 & \underline{86.5} & 93.7 & \textbf{45.2} & \textbf{74.3} & \textbf{84.5} & \underline{32.4} & \textbf{63.0} & \underline{75.0} \\
    \hline 
    \end{tabular}
    \vspace{5pt}
    \caption{\textbf{MS-COCO results.} Besides our results, we also provide previously reported results to facilitate comprehensive comparisons.}
    \label{tab:results_coco} 
    \end{table*}

    \section{Experiments}
    \label{sec:experiments}
    
    We evaluate our approach on image-text and video-text cross-modal retrieval scenarios. For image-text cross-retrieval, we evaluate on the MS-COCO dataset~\cite{lin-eccv14}; for video-text we use the TGIF~\cite{li-cvpr16} and our MRW datasets. 
    
    For MS-COCO we use the data split of \cite{kiros-arxiv14}, which provides 113,287 training, 5K validation and 5K test samples; each image comes with 5 captions. We report results on both 1K unique test images (averaged over 5 folds) and the full 5K test images. For TGIF we use the original data split~\cite{li-cvpr16} with 80K training, 10,708 validation and 34,101 test samples; since most test videos come with 3 captions, we report results on 11,360 unique test videos. For MRW, we use a data split of 44,107 training, 1K validation and 5K test samples; all the videos come with one caption. 
    
    Following the convention in cross-modal retrieval, we report results using Recall@$k$ (R@$k$) at $k=1, 5, 10$, which measures the the fraction of queries for which the correct item is retrieved among the top $k$ results. We also report the median rank (Med R) of the closest ground truth result in the list, as well as the normalized median rank (nMR) that divides the median rank by the number of total items. For cross-validation, we select the best model that achieves the highest $rsum = R@1 + R@5 + R@10$ in both directions (visual-to-text and text-to-visual) on a validation set.
    
    While we report quantitative results in the main paper, our supplementary material contains qualitative results with visualizations of multi-head self-attention maps.
    
    \subsection{Image-Text Retrieval Results}
    \label{subsec:image-to-text}
    
    Table~\ref{tab:results_coco} shows the results on MS-COCO. To facilitate comprehensive comparisons, we provide previously reported results on this dataset.\footnote{We omit results from cross-attention models~\cite{huang-cvpr17,lee-eccv18} that require a pair of instances (e.g., image and text) when encoding each instance.} Our approach outperforms most of the baselines, and achieves the new state-of-the-art on the image-to-text task on the 5K test set. We note that both GXN~\cite{gu-cvpr18} and SCO~\cite{huang-cvpr18} are trained with multiple objectives; in addition to solving the ranking task, GXN performs image-text cross-modal synthesis as part of training, while SCO performs classification of semantic concepts and their orders as part of training. Compared to the two methods, our model is trained with a single objective (ranking) and thus could be considered as a simpler model. 
    
    The most direct comparison to ours would be with VSE++~\cite{faghri-bmvc17}. Both our model and VSE++ share the same image and sentence encoders. When we let our PIE-Net to produce single embeddings for input instances (K=1), the only difference becomes that VSE++ directly uses our global features as their embedding representations, while we use the output from our PIE-Nets. The performance gap between ours (K=1) and VSE++ shows the effectiveness of our PIE-Net, which combines global context with locally-guided features produced by our local feature transformer.
    
    \subsection{Video-Text Retrieval Results}
    \label{subsec:image-to-text}
    
    Table~\ref{tab:results_tgif} and Table~\ref{tab:results_mrw} show the results on TGIF and MRW datasets. Because there is no previously reported results on these datasets for the cross-model retrieval scenario, we run the baseline models and report their results. We can see that our method show strong performance compared to all the baselines. We provide implementation details of the baseline models in the supplementary material. 
    
    We notice is that the overall performance is much lower than the results from MS-COCO. This shows how challenging video-text retrieval is (and video understanding in a broader context), and calls for further research in this task. We can also see that there is a large performance gap between the two datasets. This suggests the two datasets have significantly different characteristics: the TGIF contains sentences describing visual content in videos, while our MRW dataset contains videos showing one of possible reactions to certain situations described in sentences. This makes the association between video and text modalities much weaker for the MRW than for the TGIF.
    
    \begin{table*}[t]
    \centering
    \begin{tabular}{l|cccc|cccc}
    \hline
    \multicolumn{1}{c|}{\multirow{2}{*}{Method}} & \multicolumn{4}{c|}{Video-to-Text} & \multicolumn{4}{c}{Text-to-Video} \\ 
      & R@1 & R@5 & R@10 & Med R (nMR) & R@1 & R@5 & R@10 & Med R (nMR)\\ 
    \hline 
    VSE++~\cite{faghri-bmvc17}
        & 1.42 & 5.63 & 9.60 & 192 (0.02) & 1.55 & 5.89 & 9.77 & 220 (0.04) \\
    Order~\cite{vendrov-iclr16}
        & 1.67 & 5.49 & 9.20 & 223 (0.02) & 1.58 & 5.57 & 9.41 & 205 (0.02) \\
    Corr-AE~\cite{feng-mm14}
        & 2.15 & 7.29 & 11.47 & 158 (0.01) & 2.10 & 7.38 & 11.86 & 148 (0.01) \\
    DeViSE~\cite{frome-nips13}
        & 2.10 & 7.42 & 11.90 & 159 (0.01) & 2.19 & 7.64 & 12.52 & 146 (0.01) \\
    \hline 
    PVSE (K=1) 
        & 2.82 & 9.07 & 14.02 & 128 (0.01) & 2.63 & 9.37 & 14.58 & 115 (0.01) \\
    \textbf{PVSE}  
        & \textbf{3.28} & \textbf{9.87} & \textbf{15.56} & \textbf{115 (0.01)} & \textbf{3.01} & \textbf{9.70} & \textbf{14.85} & \textbf{109 (0.01)}  \\
    \hline 
    \end{tabular}
    \vspace{1pt}
    \caption{Experimental results on the TGIF dataset.}
    \label{tab:results_tgif} 
    \end{table*}

    \begin{table*}[t]
    \centering
    \begin{tabular}{l|cccc|cccc}
    \hline
    \multicolumn{1}{c|}{\multirow{2}{*}{Method}} & \multicolumn{4}{c|}{Video-to-Text} & \multicolumn{4}{c}{Text-to-Video} \\
      & R@1 & R@5 & R@10 & Med R (nMR) & R@1 & R@5 & R@10 & Med R (nMR) \\ 
    \hline 
    VSE++~\cite{faghri-bmvc17}
        & 0.06 & 0.18 & 0.42 & 1981 (0.40) & 0.10 & 0.34 & 0.68 & 1967 (0.39) \\
    Order~\cite{vendrov-iclr16}
        & 0.08 & 0.42 & 0.80 & 1871 (0.37) & 0.14 & 0.42 & 0.74 & 1864 (0.37) \\
    Corr-AE~\cite{feng-mm14}
        & 0.08 & 0.22 & 0.60 & 1661 (0.33) & 0.14 & 0.44 & 0.74 & 1608 (0.32) \\
    DeViSE~\cite{frome-nips13}
        & 0.08 & 0.46 & 0.86 & 1685 (0.34) & 0.10 & 0.42 & 0.74 & 1583 (0.32) \\
    \hline 
    PVSE (K=1) 
        & 0.16 & \textbf{0.68} & 0.90 & 1700 (0.34) & 0.16 & 0.56 & 0.88 & 1650 (0.33)  \\
    \textbf{PVSE} 
        & \textbf{0.18} & 0.62 & \textbf{1.18} & \textbf{1624 (0.32)} & \textbf{0.20} & \textbf{0.70} & \textbf{1.16} & \textbf{1552 (0.31)} \\
    \hline 
    \end{tabular}
    \vspace{1pt}
    \caption{Experimental results on the MRW dataset.}
    \label{tab:results_mrw} 
    \end{table*}

    \begin{figure}
       \centering
       \includegraphics[width=\linewidth]{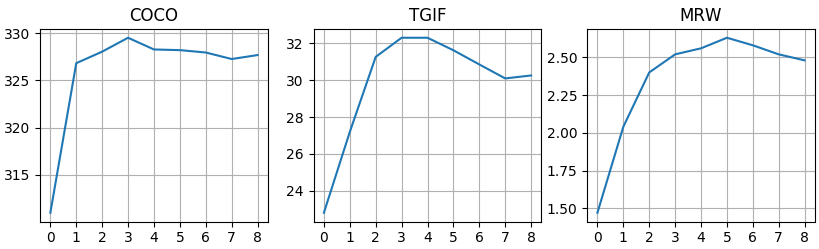}
       \caption{Performance (\textit{rsum}) with different numbers of embeddings, $K=[0:8]$. The results at $K=0$ is when we take out the PIE-Net and use the global feature as the embedding output.}
       \label{fig:ablation_num_embeds}
    \end{figure}
    
    \begin{figure}
       \centering
       \includegraphics[width=.95\linewidth]{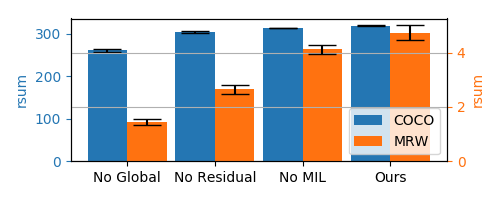}
       \caption{Performance (\textit{rsum}) on MS-COCO and MRW with different ablative settings. The error bars are obtained from multiple runs over $K=[1:8]$.}
       \label{fig:ablation_attn_mil}
    \end{figure}

    \begin{figure}
       \centering
       \includegraphics[width=0.95\linewidth]{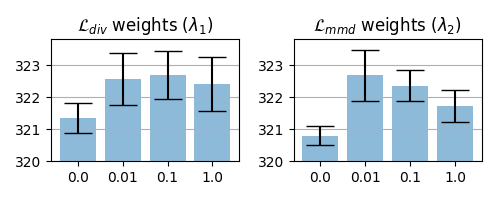}
       \caption{Performance (\textit{rsum}) on MS-COCO with different loss weights for $\mathcal{L}_{div}$ and $\mathcal{L}_{mmd}$. The error bars are obtained from multiple runs of $K=[2:4]$ and $\lambda_{(\cdot)}=[0.0, 0.01, 0.1, 1.0]$. }
       \label{fig:ablation_div_mil}
    \end{figure}

    \subsection{Ablation Results}
    \label{subsec:image-to-text}
    
    \textbf{The number of embeddings $K$:} Tables~\ref{tab:results_coco},~\ref{tab:results_tgif},~\ref{tab:results_mrw} show that computing multiple embeddings per instance improves performance compared to just a single embedding (see the last two rows in each table). To better understand the effect of $K$, we vary it from 1 to 8, and also compare with $K=0$, a baseline where we bypass our Local Feature Transformer and simply use the global feature as the final embedding representation. Figure~\ref{fig:ablation_num_embeds} shows the performance on all three datasets based on the \textit{rsum} metric (R@1 + R@5 + R@10 for image/video-to-text and back). The results are from the models before fine-tuning the ResNet-152 weights. We can see that there is a significant improvement from $K=0$ to $K=1$; this shows the effectiveness of our Local Feature Transformer. We can make an interesting observation by comparing the optimal $K$ settings across different datasets: $K=3$ for COCO and TGIF, and $K=5$ for MRW. While this cannot be used as strong evidence, we believe this shows the level of ambiguity is higher on MRW than the other two datasets. 
    
    \textbf{Global vs. locally-guided features:} We analyze the importance of global and locally-guided features, as well as different strategies to combine them. Figure~\ref{fig:ablation_attn_mil} shows results on several ablative settings: \texttt{No Global} is when we use locally-guided features alone (discard global features); \texttt{No Residual} is when we simply concatenate global and locally-guided features, instead of combining them via residual learning. We report results on both MS-COCO and MRW because the two datasets exhibit the biggest difference in the level of ambiguity. 
    
    We notice that the performance drops significantly on both datasets when we discard global features. Together with $K=0$ results in Figure~\ref{fig:ablation_num_embeds} (discard locally-guided features), this shows the importance of balancing global and local information in the final embedding. We also see that simply concatenating the two features (no residual learning) hurts the performance, and the drop is more significant on the MRW dataset. This suggests our residual learning setup is especially crucial for highly ambiguous data.  
    
    \textbf{MIL objective:} Figure~\ref{fig:ablation_attn_mil} also shows the result of \texttt{No MIL}, which is when we concatenate the $K$ embeddings and optimize the standard triplet ranking objective~\cite{frome-nips13,kiros-arxiv14,faghri-bmvc17}, i.e., the ``Conventional'' setup in Figure~\ref{fig:mil-concept}. While the differences are relatively smaller than with the other ablative settings, there are statistically significant differences between the two results on both datasets ($p=0.046$ on MS-COCO and $p=0.015$ on MRW). We also see that the difference between \texttt{No MIL} and \texttt{Ours} on MRW is more pronounced than on MS-COCO. This suggests thed MIL objective is especially effective for highly ambiguous data.  
    
    \textbf{Sensitivity analysis on different loss weights:} Figure~\ref{fig:ablation_div_mil} shows the sensitivity of our approach when we vary the relative loss weights, i.e., $\lambda_1$ and $\lambda_2$ in Equation~\eqref{eq:our_objective}. Note that the weights are relative, not absolute, e.g., instead of directly multiplying $\lambda_1 = 1.0$ to $\mathcal{L}_{div}$, we first scale it to $\lambda_1 \times (\mathcal{L}_{mil} / \mathcal{L}_{div})$ and then multiply it to $\mathcal{L}_{div}$. The results show that both loss terms are important in our model. We can see, in particular, that $\mathcal{L}_{mmd}$ plays an important role in our model. Without it, the two embedding spaces induced by different modalities may diverge quickly due to the MIL objective, which may result in a poor convergence rate. Overall, our results suggests that the model is not much sensitive to the two relative weight terms.


    
    
    \section{Conclusion}
    \label{sec:conclusion}
    Ambiguous instances and their partial associations pose significant challenges to cross-modal retrieval. Unlike the traditional approaches that use injective embedding to compute a single representation per instance, we propose a Polysemous Instance Embedding Network (PIE-Net) that computes \textit{multiple and diverse} representations per instance. To obtain visual-semantic embedding that is robust to partial cross-modal association, we tie-up two PIE-Nets, one per modality, and jointly train them using the Multiple Instance Learning objective. We demonstrate our approach on the image-text and video-text cross-modal retrieval scenarios and report strong results compared to several baselines. 
    
    Part of our contribution is also in the newly collected MRW dataset. Unlike existing video-sentence datasets that contain sentences describing visual content in videos, ours contain videos illustrating \textit{one of possible reactions} to certain situations described in sentences, which makes video-sentence association somewhat ambiguous. This poses new challenges to cross-modal retrieval; we hope there will be further progress on this challenging new dataset. 
    
    \newpage
    
    \appendix
    \appendixpage
    
    \section{MRW Dataset}
    \label{sec:dataset}
    
    Our dataset consists of 50,107 video-sentence pairs collected from popular social media websites including reddit, Imgur, and Tumblr. We crawled the data using the GIPHY API\footnote{\url{https://developers.giphy.com}} with query terms \texttt{mrw}, \texttt{mfw}, \texttt{hifw}, \texttt{reaction}, and \texttt{reactiongif}; we crawled the data from August 2016 to March 2019. Table~\ref{tab:descriptive_statistics} shows the descriptive statistics of our dataset. We are continuously crawling the data, and plan to release updated versions in the future.

    \begin{table*}[t]
    \centering
    \begin{tabular}{l||ccc|c}
    & Train & Validation & Test & Total \\ \hline\hline
    Number of video-sentence pairs & 44,107 & 1,000 & 5,000 & 50,107 \\ \hline
    Average / median number of frames & 104.91 / 72 & 209.04 / 179 & 209.55 / 178 & 117.43 / 79\\
    Average / median number of words & 11.36 / 10 & 15.02 / 14 & 14.79 / 13 & 11.78 / 11 \\ 
    Average / median word frequency & 15.48 / 1 & 4.80 / 1 & 8.57 / 1 & 16.94 / 1\\ \hline
    Vocabulary size & 34,835 & 34,835 & 34,835 & 34,835 \\ \hline
    \end{tabular}
    \caption{Descriptive statistics of the MRW dataset.}
    \label{tab:descriptive_statistics} 
    \end{table*}

    \subsection{Previous Work on Animated GIF}
    
    Note that most of the videos in our dataset have the animated GIF format. Technically speaking, animated GIFs and videos have different formats; the former is lossless, palette-based, and has no audio. In this paper, however, we use the two terms interchangeably because the distinction is unnecessary in our method. Below, to provide the context for our work, we briefly review previous work that focused on animated GIF.
    
    There is increasing interest in conducting research around animated GIFs. Bakhshi~\etal~\cite{bakhshi-chi16} studied what makes animated GIFs engaging on social networks and identified a number of factors that contribute to it: the animation, lack of sound, immediacy of consumption, low bandwidth and minimal time demands, the storytelling capabilities and utility for expressing emotions. Previous work in the computer vision and multimedia communities used animated GIFs for various tasks in video understanding. Jou~\etal~\cite{jou-mm14} propose a method to predict viewer perceived emotions for animated GIFs. Gygli~\etal~\cite{gygli-cvpr16} propose the Video2GIF dataset for video highlighting, and further extended it to emotion recognition~\cite{gygli-mm16}. Chen~\etal~\cite{chen-acii17} propose the GIFGIF+ dataset for emotion recognition. Zhou~\etal~\cite{zhou-wacv18} propose the Image2GIF dataset for video prediction, along with a method to generate cinemagraphs from a single image by predicting future frames. 
    
    Recent work use animated GIFs to tackle the vision \& language problems. Li~\etal~\cite{li-cvpr16} propose the TGIF dataset for video captioning; Jang~\etal~\cite{jang-cvpr17} propose the TGIF-QA dataset for video visual question answering. Similar to the TGIF dataset~\cite{li-cvpr16}, our dataset includes video-sentence pairs. However, our sentences are created by real users from Internet communities rather than study participants, thus posing real-world challenges. More importantly, our dataset has \textit{implicit} concept association between videos and sentences (videos contain physical or emotional reactions to sentences), while the TGIF dataset has \textit{explicit} concept association (sentences describe visual content in videos).

    \subsection{Analysis of Facial Expressions}
    \label{subsec:data_analysis}
    \begin{figure*}
       \includegraphics[trim={70pt 0pt 80pt 0pt},clip,width=\linewidth]{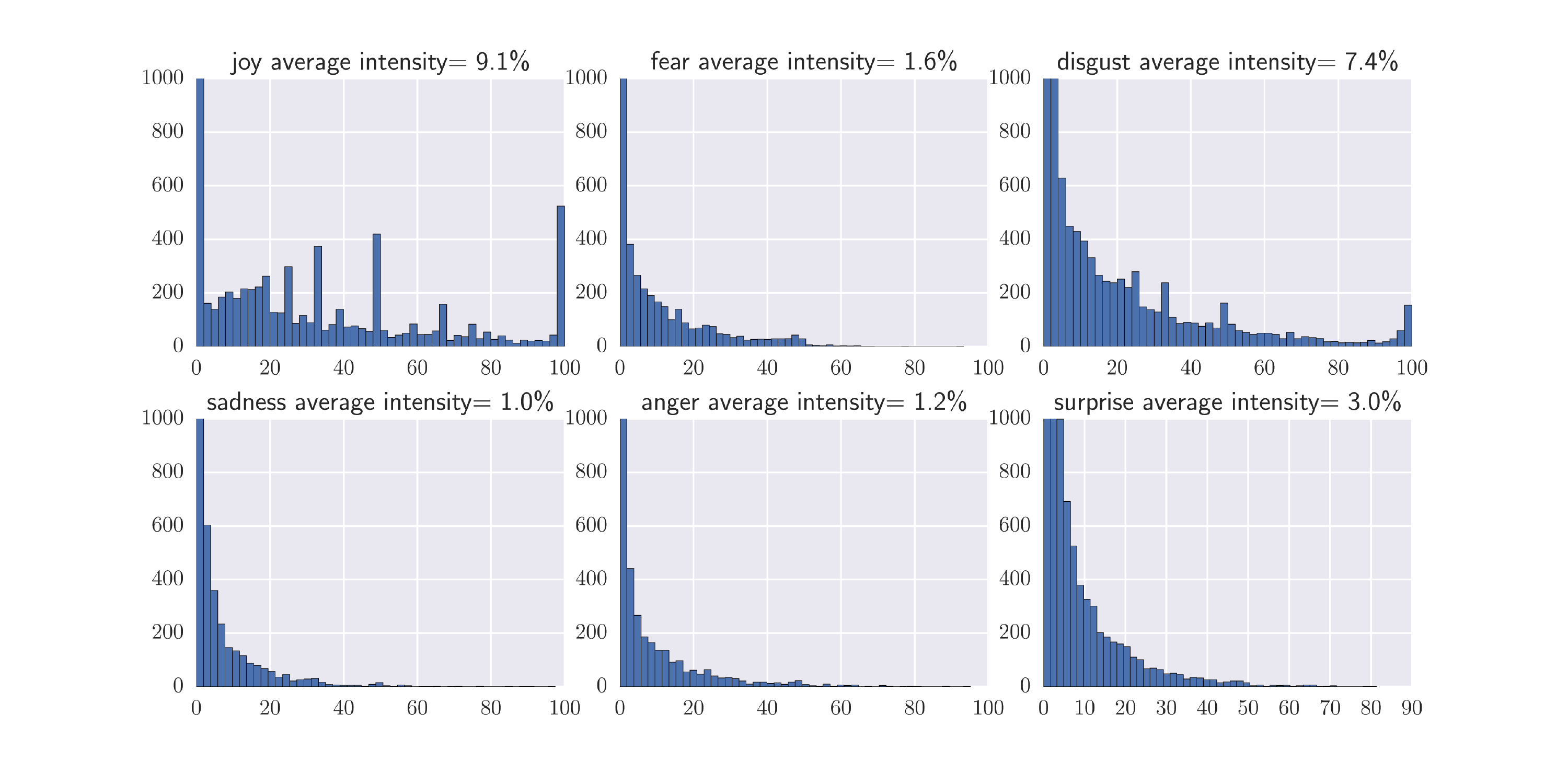}
       \caption{Histograms of the intensity of facial expressions. The horizontal axis represents the intensity of the detected expression, while the vertical axis is the sample count in frames with faces. We clip the $y$-axis at 1000 for visualization. Overall, joy, with average intensity of 9.1\% and disgust (7.4\%) are the most common facial expressions in our dataset.} 
       \label{fig:expression}
    \end{figure*}
    Facial expression plays an important role in our dataset: 6,380 samples contain the hashtag MFW (my face when), indicating that those GIFs contain emotional reactions manifested by facial expressions. To better understand the landscape of our dataset, we analyze the types of facial expressions contained in our dataset by leverage automatic tools. 
    
    First, we count the number of faces appearing in the animated GIFs. To do this, we applied the dlib CNN face detector~\cite{dlib09} on five frames sampled from each animated GIF with an equal interval. The results show that there are, on average, $0.73$ faces in a given frame of an animated GIF. Also, 34,052 animated GIFs contain at least one face. This means that 72\% of our videos contain faces, which is quite significant. This suggests that employing techniques tailored specifically for face understanding could potentially improve performance on our dataset.
    
    Next, we use the Affectiva Affdex~\cite{McDuff2016Affdex} to analyze facial expressions depicted in the animated GIFs, detecting the intensity of expressions from two frames per second in each animated GIF. We looked at six expressions of basic emotions~\cite{ekman1992argument}, namely, joy, fear, sadness, disgust, surprise and anger. We analyzed only the frames that contain a face with its bounding box region larger than 15\% of the image. Figure~\ref{fig:expression} shows the results. Overall, joy with average intensity of 9.1\% and disgust (7.4\%) are the most common facial expressions in our dataset.

    \subsection{Comparison to the TGIF Dataset}
    \label{subsec:data_analysis}
    
    Image and video captioning often involves describing objects and actions depicted explicitly in visual content~\cite{lin-eccv14,li-cvpr16}. For reaction GIFs, however, visual-textual association is not always explicit. For example, as is the case in our dataset, objects and actions depicted in visual content might be a physical or emotional reaction to the scenario posed in the sentence. 
    
    In this section, we qualitatively compare our dataset with the TGIF dataset~\cite{li-cvpr16}, which contains 120K video-sentence pairs for video captioning. We chose the dataset because both datasets contain animated GIFs collected from social media, and thus contain similar visual content.
    
    \begin{figure*}
       \centering
       \includegraphics[width=\linewidth]{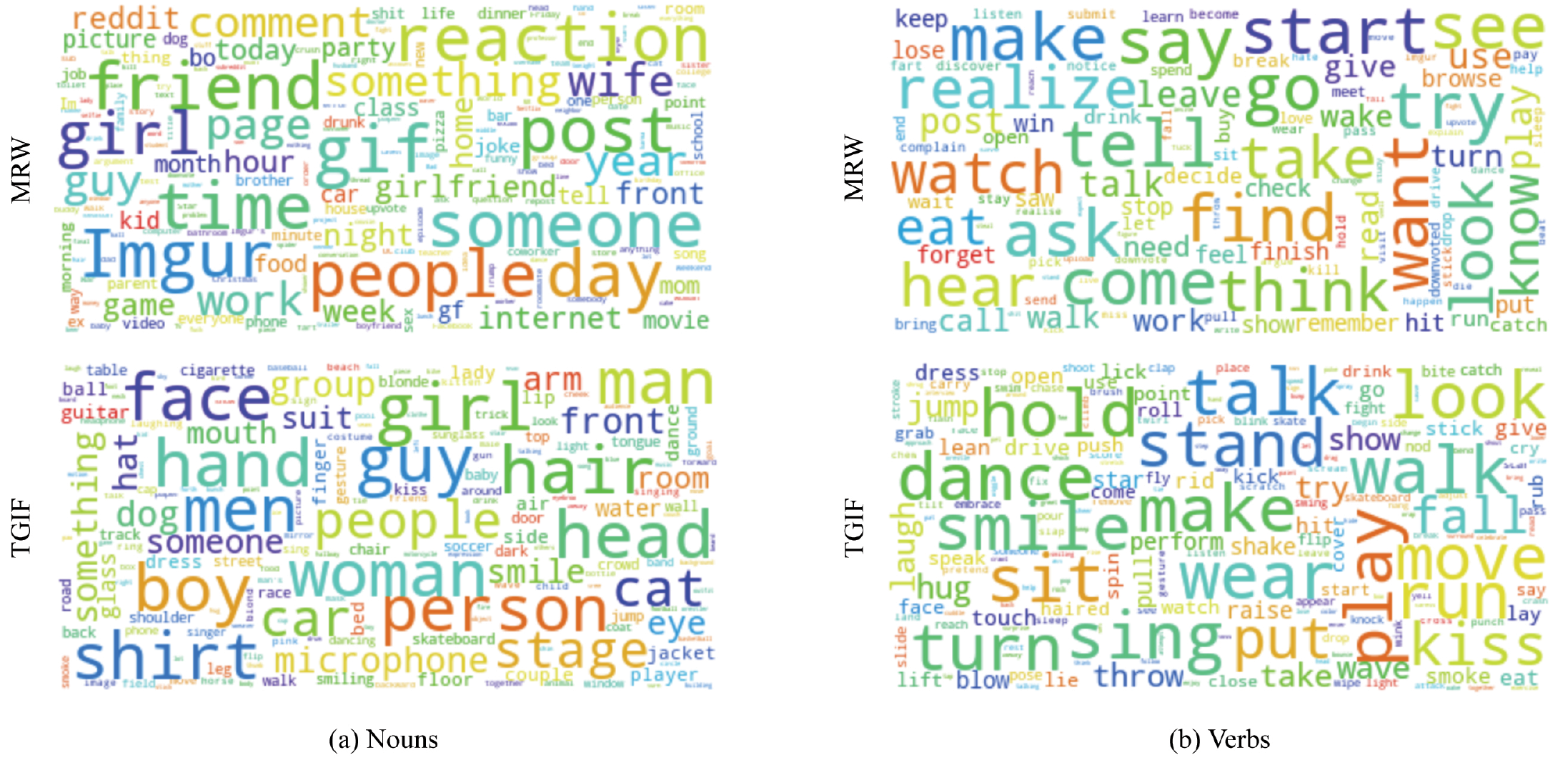}
       \caption{Distributions of nouns and verbs in our MRW and the TGIF~\cite{li-cvpr16} datasets. Compared to the TGIF dataset, words in our dataset depict more abstract concepts (e.g., post, time, day, start, realize, think, try), suggesting the ambiguous nature in our dataset.}
       \label{fig:wordcloud}
    \end{figure*}
    We first compare words appearing in both datasets. Figure~\ref{fig:wordcloud} shows word clouds of nouns and verbs extracted from our MRW dataset and the TGIF dataset~\cite{li-cvpr16}. Sentences in the TGIF dataset are constructed by crowdworkers to describe the visual content explicitly displayed in animated GIFs. Therefore, its nouns and verbs mainly describe physical objects, people and actions that can be visualized, e.g., cat, shirt, stand, dance. In contrast, MRW sentences are constructed by the Internet users, typically from subcommunities in social networks that focus on reaction GIFs. As can be seen from Figure~\ref{fig:wordcloud}, verbs and nouns in our MRW dataset additionally include abstract terms that cannot necessarily be visualized, e.g., time, day, realize, think. This shows that our dataset contains ambiguous terms and their associations, which pose significant challenges to cross-modal retrieval.
    
    Next, we compare whether video-sentence associations are explicit/implicit in both datasets. To this end, we conducted a user study in which we asked six participants to verify the association between sentences and animated GIFs. We randomly sampled 100 animated GIFs from the test sets of both our dataset and TGIF dataset~\cite{li-cvpr16}. We paired each animated GIF with both its associated sentence and a randomly selected sentence from the corresponding dataset, resulting in 200 GIF-sentence pairs per dataset. 
    
    The results show that, in case of our dataset (MRW), 80.4\% of the associated pairs are positively marked as being relevant, suggesting humans are able to distinguish the true vs. fake pairs despite implicit concept association. On the other hand, 50.7\% of the randomly assigned sentences are also marked as matching sentences. The high false positive rate shows the ambiguous nature of GIF-sentence association in our dataset. 
    
    In contrast, for the TGIF dataset with clear explicit association, 95.2\% of the positive pairs are correctly marked as relevant and only 2.6\% of the irrelevant pairs are marked as being relevant. This human baseline demonstrates the challenging nature of GIF-sentence association in our dataset, due to their implicit rather than explicit association.

    \subsection{Application: Animated GIF Search}
    Animated GIFs are becoming increasingly popular~\cite{bakhshi-chi16}; more people use them to tell stories, summarize events, express emotion, and enhance (or even replace) text-based communication. To reflect this trend, several social networks and messaging apps have recently incorporated GIF-related features into their systems, e.g., Facebook users can create posts and leave comments using GIFs, Instagram and Snapchat users can put ``GIF stickers'' into their personal videos, and Slack users can send messages using GIFs. This rapid increase in popularity and real-world demand necessitates more advanced and specialized systems for animated GIF search. 
    
    Current solutions to animated GIF search rely entirely on concept tags associated with animated GIFs and matching them with user queries. The tags are typically provided by users or produced by editors at companies like GIPHY. In the former case, noise becomes an issue; in the latter, it is expensive and would not scale well.   
    
    One of the motivations behind collecting our MRW dataset is to build a text-based animated GIF search engine, targeted for real-world scenarios mentioned above. Existing video captioning datasets, such as TGIF~\cite{li-cvpr16}, are inappropriate for our purpose because of the explicit nature of visual-textual association, i.e., sentences simply describe what is being shown in videos. Rather, we need a dataset that captures various types of nuances used in social media, e.g., humor, irony, satire, sarcasm, incongruity, etc. Because our dataset provides video-text pairs with implicit visual-textual association, we believe that it has the potential to provide training data for building text-based animated GIF search engines targeted for social media.
    
    To demonstrate the potential, we provide qualitative results on text-to-video retrieval using our dataset, shown in Figure~\ref{fig:mrw_attn}. Each set of results show a query text and the top five retrieved videos, along with their ranks and cosine similarity scores. We would like the readers to take a close look at each set of results and decide which of the five retrieved videos depict the most likely visual response to the query sentence. The answers are provided below. For better viewing experience, we provide an HTML page with animated GIFs instead of static images. We strongly encourage the readers to check the HTML page to better appreciate the results. \textit{\scriptsize (Answers: 3, 5, 2, 4, 1, 5, 4)}

    \begin{table*}[t]
    \centering
    \begin{equation}
        \mathcal{L}_{DeViSE} = \frac{1}{N} \sum_{i,j,k=1}^{N} 
        \max\left(0, \rho - d(\phi(x_i), \phi(y_i)) + d(\phi(x_j), \phi(y_k) \right), \:\: \forall (i=j \lor i=k) \land j \neq k
    \label{eq:devise}
    \end{equation} \\
    \begin{equation}
        \mathcal{L}_{VSE++} = \frac{1}{N} \sum_{i=1}^{N} \sum_{q=\{j, k\}} \max_{q}
        \max\left(0, \rho - d(\phi(x_i), \phi(y_i)) + d(\phi(x_j), \phi(y_k) \right), \:\: \forall (i=j \lor i=k) \land j \neq k
    \label{eq:vsepp}
    \end{equation}
    \begin{equation}
        \mathcal{L}_{CorrAE} = \mathcal{L}_{DeViSE} + 
        \frac{1}{N} \sum_{i=1}^N \sum_{c_i=\{x_i, y_i\}}\left( 
            \| \phi(x_i) - \tilde{\phi}(x_i|c_i)\|_2^2 +
            \| \phi(y_i) - \tilde{\phi}(x_i|c_i)\|_2^2 
        \right)
    \label{eq:corrae}
    \end{equation}
    \begin{tabular}{l}
    \end{tabular}
    \label{tab:equations} 
    \end{table*}
    
    \section{Baseline Implementation Details}
    \label{sec:baseline}
    In the experiment section, we provided baseline results for MS-COCO, TGIF, and MRW datasets. For MS-COCO, we provided previously reported results. For TGIF and MRW, on the other hand, we reported our own results because there has not been previous results on the datasets. Due to the space limit, we omitted implementation details of the baseline approaches; here we provide implementation details of the four baseline approaches: DeViSE~\cite{frome-nips13}, VSE++~\cite{faghri-bmvc17}, Order Embedding~\cite{vendrov-iclr16}, and Corr-AE~\cite{feng-mm14}.
    
    For fair comparison, all four baselines share the same video and sentence encoders as described in Section 3.1 of the main paper. The only difference is in the loss function we train the models with. Following the notation used in the main paper, we denote the output of the video and sentence encoders by $\phi(x)$ and $\phi(y)$, respectively. We employ the following loss functions for the baselines:
    
    \textbf{DeViSE~\cite{frome-nips13}:} We implement the conventional hinge loss in the triplet ranking setup; see Equation~\eqref{eq:devise}. It penalizes the cases when the distance between positive pairs (i.e., the ground truth) is further away than negative pairs (e.g., randomly sampled) with a margin parameter $\rho$ (we measure the cosine distance).
    
    \textbf{VSE++~\cite{faghri-bmvc17}:} We implement the hard negative mining version of the conventional hinge loss triplet ranking loss; see Equation~\eqref{eq:vsepp}. We have experimented with the original version and found that it fails to find a suitable solution to the objective, producing retrieval results that are almost identical to random guess. We suspect that the high noise present in both TGIF and MRW datasets makes the max function too strict as a constraint. We therefore replace the $\max_q$ function with a ``filter'' function that includes only highly-violating cases while ignoring others. 
    
    Intuitively, we implement the filter function to be an outlier detection function based on z-scores, where any z-score greater than 3 or less than -3 is considered to be an outlier. Specifically, we compute the z-scores for all of possible $(i,j,k)$ combinations inside Equation~\eqref{eq:vsepp} and discard instances if their absolute z-score is below 3.0. This way, we are considering multiple hard negatives instead of just one. We have empirically found this modification to be crucial to achieve reasonable performances on the TGIF and MRW datasets.
    
    \textbf{Order Embedding~\cite{vendrov-iclr16}:} We used the original implementation provided by the authors of \cite{vendrov-iclr16}.
    
    \textbf{Corr-AE~\cite{feng-mm14}:} We implement the correspondence cross-modal autoencoder proposed by Feng~\etal~\cite{feng-mm14} (see Figure 4 in \cite{feng-mm14}). Given the encoder output $\phi(x)$ and $\phi(y)$, we build two autoencoders, one per modality, so that each autoencoder can reconstruct both $\phi(x)$ and $\phi(y)$. The autoencoders have four fully-connected layers with [512, 256, 256, 512] hidden units, respectively. Each of the fully connected layers is followed by a ReLU activation and a layer normalization~\cite{ba-arxiv16}. 
    
    Formally, a video autoencoder takes as input $\phi(x)$ and outputs $[\tilde{\phi}(x|x); \tilde{\phi}(y|x)]$, and a sentence autoencoder takes as input $\phi(y)$ and outputs $[\tilde{\phi}(x|y); \tilde{\phi}(y|y)]$. We then train the model by optimizing the loss form shown in Equation~\eqref{eq:corrae}. We note that this loss is different from the original formulation of Corr-AE~\cite{feng-mm14}, where the first term in Equation~\eqref{eq:corrae} is replaced by a Euclidean loss, i.e., $\mathcal{L}_2 = \frac{1}{N}\sum_{i=1}^N \left( \| \phi(x_i) - \phi(y_i) \|_2^2\right)$. We found that using $\mathcal{L}_2$ instead of $\mathcal{L}_{DeViSE}$ makes the learning much harder, producing results that is almost identical to random guess.

    \section{Visualization of Multi-Head Self-Attention}
    \label{sec:attention}
    
    \subsection{Image-to-Text Retrieval Results on MS-COCO}
    \begin{figure*}[p]
       \centering
       \vspace{2pt}
       \includegraphics[width=\linewidth]{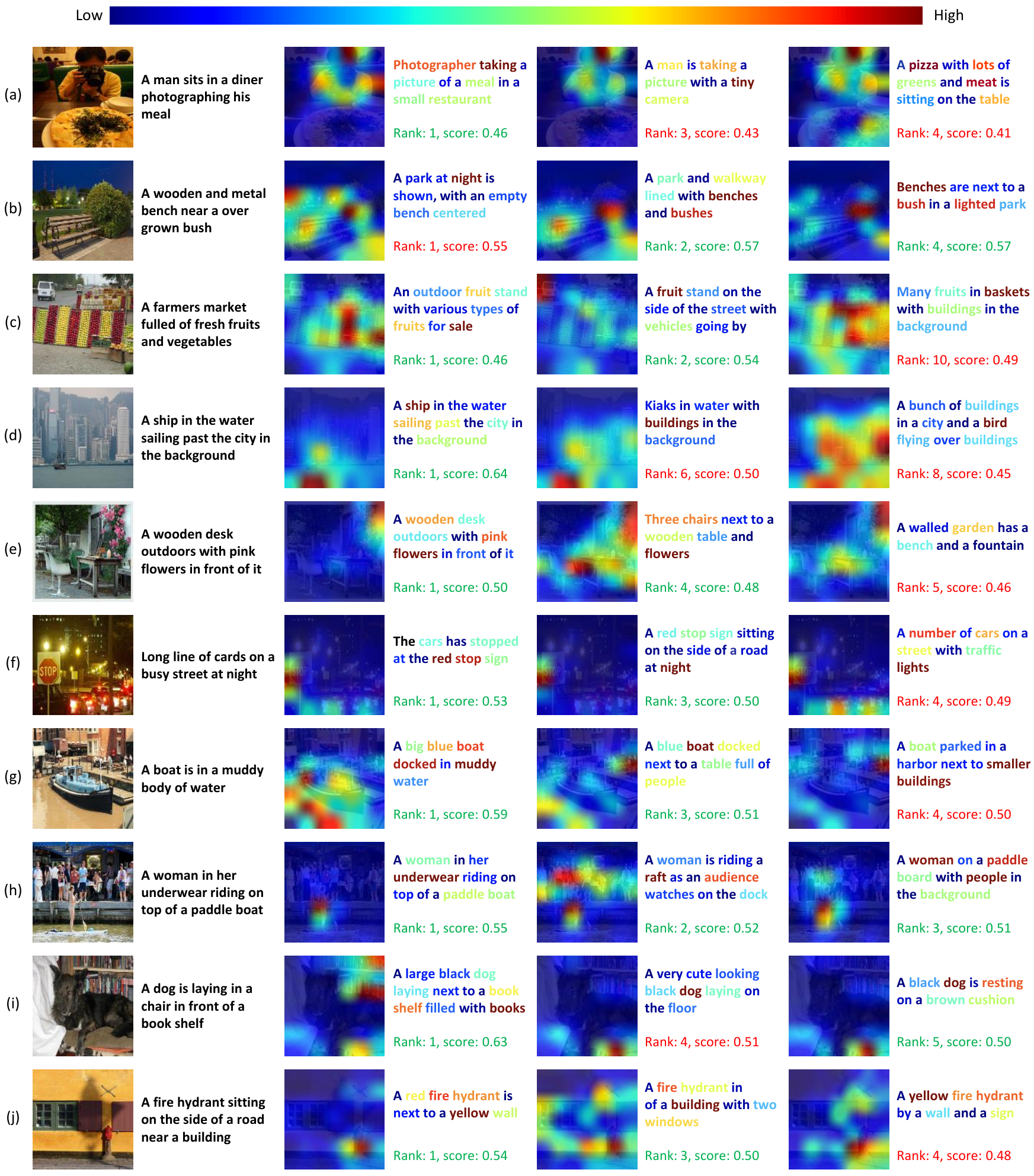}
       \vspace{2pt}
       \caption{\textbf{Image-to-text retrieval results on MS-COCO.} For each query image we show three visual attention maps and their top-ranked text retrieval results, along with their ranks and cosine similarity scores (green: correct, red: incorrect). Words in each sentence is color-coded with textual attention intensity, using the color map shown at the top. }
       \label{fig:coco_attn}
    \end{figure*}
    Figure~\ref{fig:coco_attn} shows examples of visual-textual attention maps on the MS-COCO dataset; the task is image-to-text retrieval. The first column shows query images with ground-truth sentences. Each of the other three columns shows visual (spatial) attention maps and their top-ranked text retrieval results, as well as their ranks and cosine similarity scores (green: correct, red: incorrect). We color-code words in the retrieved sentences according to their textual attention intensity values, normalized between [0, 1]. 
    
    A glimpse at the results in each row shows that the three attention maps attend to different regions of the query image. Looking closely, we notice that salient regions are typically attended by multiple attention maps. For example, all three attention maps in Figure~\ref{fig:coco_attn} highlight: (a) the photographer, (b) the bench, (c) the fruit stand, (e) the pink flowers, (f) the stop sign, (h) the woman, (j) the fire hydrant. However, this is not always the case: In Figure~\ref{fig:coco_attn} (i), none of the attention maps highlights the most salient object, the black dog, and each attention map highlights different regions in the image. Even though all three attention maps do not ``attend to'' the dog, their top-ranked text retrieval results are still highly relevant to the query image; all three retrieved sentences have the word \textit{dog} in them. This is possible because our PIE-Net computes embedding vectors by combining global context with locally-guided features. In this example, the global context provides information about the black dog, while each of the three locally-guided features contains region-specific information, specifically, (first map): the book shelf, (second map): the floor, (third map): the brown cushion.
    
    The most interesting observation is that there are subtle variations in the retrieved sentences depending on where the visual attention is focused on. For example, in Figure~\ref{fig:coco_attn} (a), the first result focuses on the photographer as a whole, the second focuses on the tiny camera (the visual attention is more narrowly focused on the photographer), and the third focuses on the pizza on the table (notice the visual attention on the table). In Figure~\ref{fig:coco_attn} (d), the first result focuses on the ship, the second focuses on the building, and the third on an (imaginary) bird that could have been flying over the buildings. In Figure~\ref{fig:coco_attn} (g), the first result focuses on the boat and the muddy water (notice visual attention on the muddy water region at the lower left corner), while the second focuses on the table of people (notice visual attention on the table region). In Figure~\ref{fig:coco_attn} (j), the first results focuses on the fire hydrant and the yellow wall that is right behind the hydrant, while the second focuses on the hydrant as well as the building with two windows (notice now the visual attention is more widely spread out than the first result). We encourage the readers to look closely at Figure~\ref{fig:coco_attn} to appreciate the subtle variations in the retrieved sentences depending on their corresponding visual attention.

    \subsection{Video-to-Text Retrieval Results on TGIF}
    \begin{figure*}[p]
       \centering
       \vspace{2pt}
       \includegraphics[width=\linewidth]{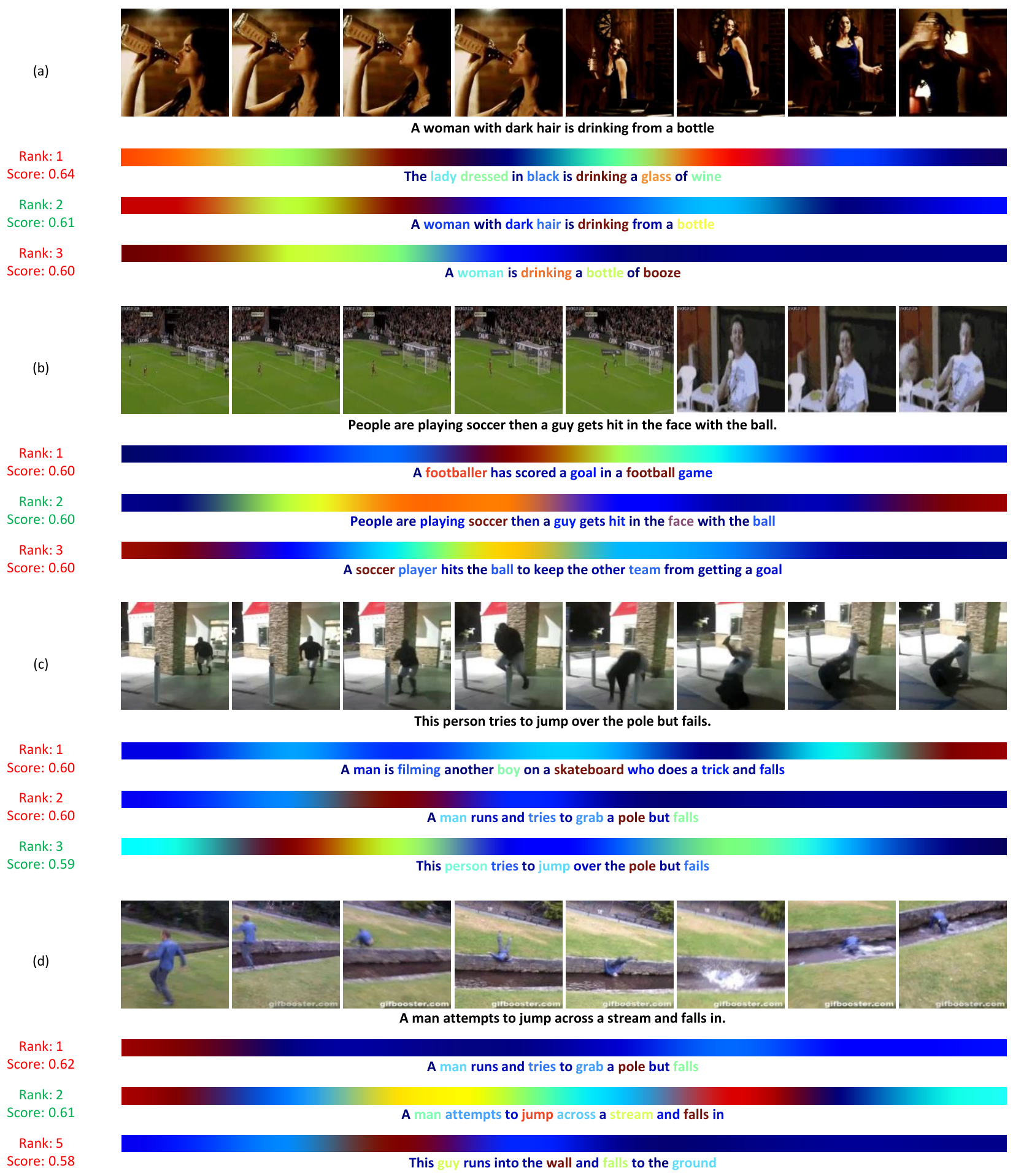}
       \vspace{2pt}
       \caption{\textbf{Video-to-text retrieval results on TGIF.} For each query video we show three visual attention maps and their top-ranked text retrieval results, along with their ranks and cosine similarity scores (green: correct, red: incorrect). Words in each sentence is color-coded with textual attention intensity.}
       \label{fig:tgif_attn}
    \end{figure*}
    Figure~\ref{fig:tgif_attn} shows examples of visual-textual attention maps on the TGIF dataset; the task is video-to-text retrieval. In each set of results, we show: (top) a query video and its ground-truth sentence, (bottom three rows): three visual (temporal) attention maps and their top-ranked text retrieval results, as well as their ranks and cosine similarity scores (green: correct, red: incorrect). We color-code words in the retrieval results according to their textual attention intensity values, normalized between [0, 1].
    
    Similar to the results on MS-COCO, here we see that visual and textual attention maps tend to highlight salient video frames and words, respectively. Looking closely, we notice that the retrieved results tend to capture the concepts highlighted by their corresponding visual attention. For example, in \textbf{Figure~\ref{fig:tgif_attn} (a)}, the top ranked result contain ``lady dressed in black'' and ''drinking a glass of wine'', and the visual attention highlights both the early part of the video, where a woman is drinking from a bottle of whisky, and the latter part, where her black dress is shown. For the second ranked result, the visual attention no longer highlights the latter part, and the retrieved text focuses solely on drinking action (no mention of her black dress). In \textbf{Figure~\ref{fig:tgif_attn} (b)}, the top ranked result focuses on scoring a goal, while the second rank result also focus on the guy being hit in the face with the ball. Notice the difference of visual attention maps between the first and the second case.

    \subsection{Text-to-Video Retrieval Results on MRW}
    \begin{figure*}[p]
       \centering
       \includegraphics[width=\linewidth]{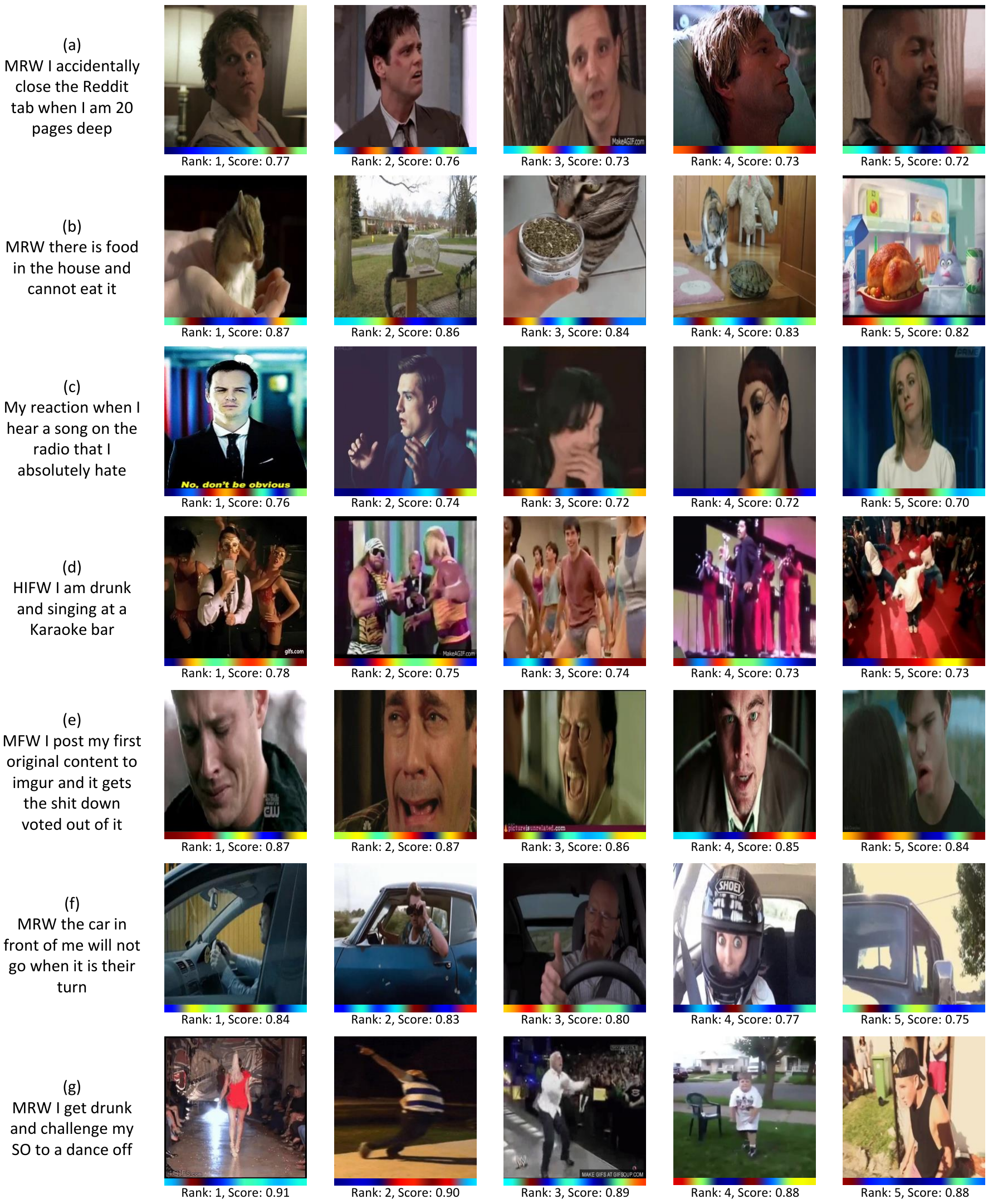}
       \caption{\textbf{Text-to-video retrieval results on MRW.} For each query sentence we show top five retrieved videos, along with their visual (temporal) attention maps, rank, and cosine similarity scores. For better viewing, we provide an HTML file with animated GIFs instead of static images. \textit{Quiz:} We encourage the readers to find the best matching video in each set of results (see the text for answers). }
       \label{fig:mrw_attn}
    \end{figure*}
    Figure~\ref{fig:mrw_attn} shows examples of text-to-video retrieval results on the MRW dataset. In each row, we show a query sentence and top five retrieved videos along with their ranks and cosine similarity scores. Unlike the previous two figures, here we do not directly show the ground-truth matches (but rather ask the readers to find them; we provide the answers above). The purpose of this is to emphasize the ambiguous and implicit nature of visual-textual association present in our dataset. 
    
    Most of the top five retrieved videos seem to be a good match to the query sentence. For example, Figure~\ref{fig:mrw_attn} (a) shows five videos that all contain a human face, each expressing subtly different emotions. Figure~\ref{fig:mrw_attn} (b) shows five videos that all contain an animal (squirrel, cat, etc), and most videos contain food. All five retrieved videos in Figure~\ref{fig:mrw_attn} show some form of awkward (dancing) moves. 
    
    We believe that the relatively poor retrieval performance reported in our main paper is partly explained by our qualitative results: visual-textual associations are highly ambiguous and there could be multiple correct matches. This calls for a different metric that measures the perceptual similarity between queries and retrieved results, rather than exact match. There has been some progress on perceptual metrics in the image synthesis literature (e.g., Inception Score~\cite{salimans-nips16}). We are not aware of a suitable perceptual metric for cross-modal retrieval, and this could be a promising direction for future research. 
    
    {\small
    \bibliographystyle{ieee_fullname}
    \bibliography{main}
    }
    
    \end{document}